\documentclass{article}
\usepackage{ijcai22}

\usepackage{times}
\usepackage{soul}
\usepackage{url}
\usepackage[hidelinks]{hyperref}
\usepackage[utf8]{inputenc}
\usepackage[small]{caption}
\usepackage{graphicx}
\usepackage{amsmath}
\usepackage{amsthm}
\usepackage{amssymb}
\usepackage{booktabs}
\usepackage{algorithm}
\usepackage{algorithmic}
\usepackage{multirow}
\usepackage{xcolor}
\usepackage{footmisc}
\usepackage{pifont}% http://ctan.org/pkg/pifont
\usepackage[export]{adjustbox}
\newcommand{\xmark}{\ding{53}}%
\usepackage{pdfrender}
\usepackage{xr}
\usepackage[titletoc]{appendix}
\newcommand*{\boldcheckmark}{%
  \textpdfrender{
    TextRenderingMode=FillStroke,
    LineWidth=.5pt, % half of the line width is outside the normal glyph
  }{\checkmark}%
}
\newcommand{\printfnsymbol}[1]{%
  \textsuperscript{\@fnsymbol{#1}}%
}
\makeatletter
\newcommand\footnoteref[1]{\protected@xdef\@thefnmark{\ref{#1}}\@footnotemark}
\makeatother

\makeatother
\title{Emotion-Controllable Generalized Talking Face Generation}

\author{
Sanjana Sinha$^1$\thanks{Equal contribution}
\and
Sandika Biswas$^1$\footnotemark[1] \and
Ravindra Yadav$^2$\thanks{Former intern at TCS Research} \And
Brojeshwar Bhowmick$^1$
\affiliations
$^1$TCS Research, India \\
$^2$IIT Kanpur, India\\
\emails
\{sanjana.sinha, biswas.sandika, b.bhowmick\}@tcs.com,
 ravin@iitk.ac.in 
}
\begin{document}

\maketitle

\begin{abstract}
    Despite the significant progress in recent years, very few of the AI-based talking face generation methods attempt to render natural emotions. Moreover, the scope of the methods is majorly limited to the characteristics of the training dataset, hence they fail to generalize to arbitrary unseen faces. In this paper,  we propose a one-shot facial geometry-aware emotional talking face generation method that can generalize to arbitrary faces. We propose a graph convolutional neural network that uses speech content feature, along with an independent emotion input to generate emotion and speech-induced motion on facial geometry-aware landmark representation.  This representation is further used in our optical flow-guided texture generation network for producing the texture. 
We propose a two-branch texture generation network, with motion and texture branches designed to consider the motion and texture content independently. 
Compared to the previous emotion talking face methods, our method can adapt to arbitrary faces captured in-the-wild by fine-tuning with only a single image of the target identity in neutral emotion.	
\end{abstract}

\section{Introduction}

Audio-driven realistic talking face generation is a widely studied research problem, with diverse applications in animation, virtual assistant, telepresence, gaming etc. Most of the existing methods \cite{chung2017you,suwajanakorn2017synthesizing,chen2019hierarchical,dasspeech,zhou2019talking,sinha2020identity,chen2020talking,zhou2020makelttalk,zhou2021pose,zhang2021flow}  mainly focus on generating realistic lip synchronization, identity preservation, eye blinks or head motion in the synthesized talking face video. 
Very few of these methods can render realistic facial emotions (Table \ref{tab:list_soa}), due to the limited availability of annotated emotional audio-visual datasets. Some earlier methods \cite{vougioukas2019realistic,chen2020talking} have tried to learn the facial emotions implicitly from the audio. However, these methods fail to control the facial emotion and often fail to produce realistic animation.
\begin{figure}[t]
    \centering
    \includegraphics[width=\columnwidth]{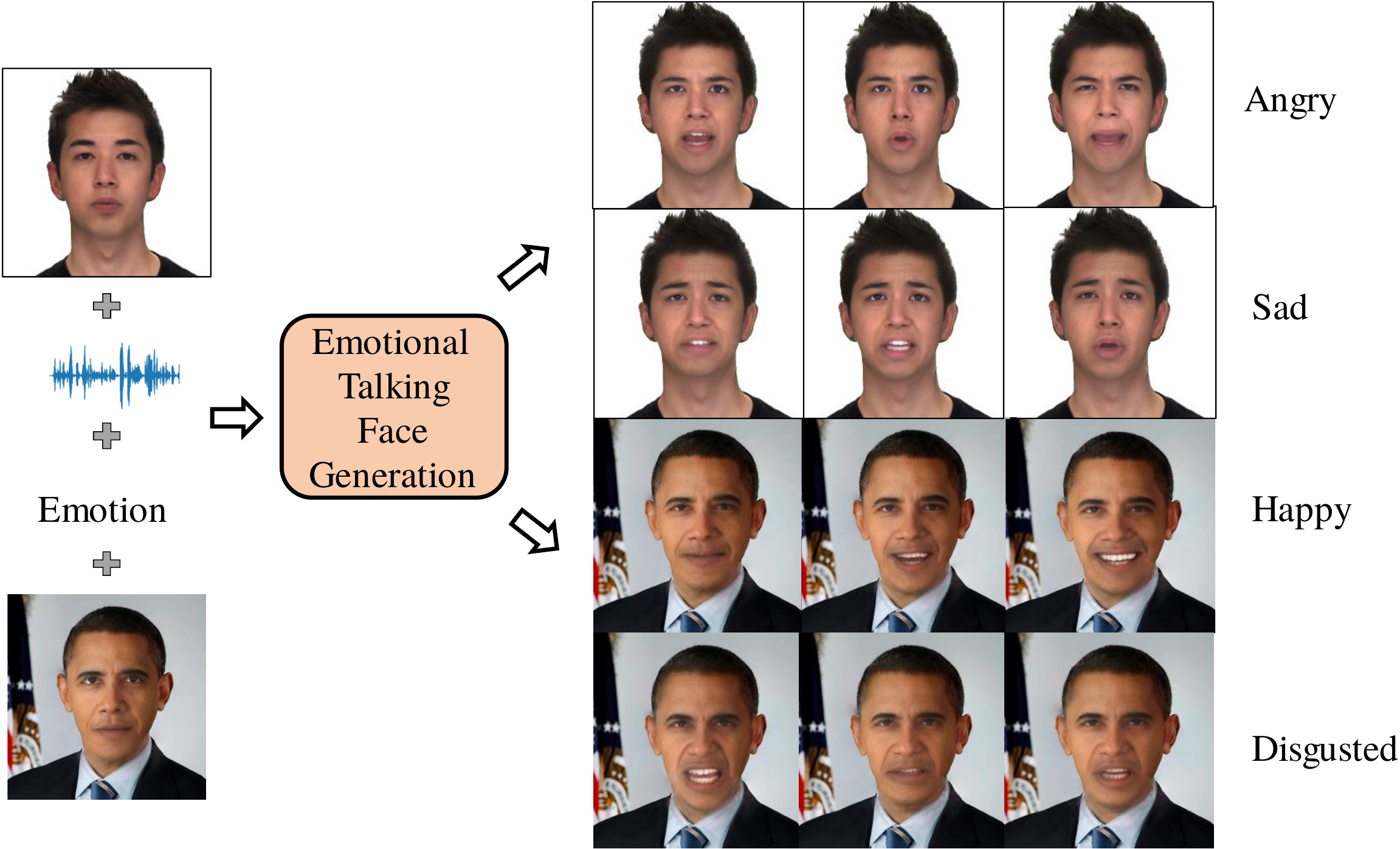}
    \caption{\small{Results of our proposed emotional talking face generation method on arbitrary faces.}}
    \label{fig:intro_img}
\end{figure}

Recently, MEAD \cite{wang2020mead} has proposed a method for emotional talking face generation with explicit emotion control and released the MEAD dataset \cite{wang2020mead} containing well-defined emotions at varying intensities, and a wide variety of sentences.  This method \cite{wang2020mead} generates emotion only in the upper face (from external emotion control using one-hot emotion vector) and the lower part of the face is animated from audio independently, which results in inconsistent emotions over the face. A recent video editing method EVP \cite{ji2021audio} focuses on generating consistent emotions over the entire face using a disentangled emotion latent feature learned from the audio. However, all these methods rely on intermediate global landmarks (or edge maps) to generate the   texture directly with emotions. To generalize the texture deformation for any unknown face for a given emotion, it is important to learn the relationship between the facial geometry and the emotion-induced local deformations within the face. None of these methods consider learning this relationship, 
%in the texture generation method, 
hence show a limited scope of generalization to an arbitrary unknown target face 
%(\textbf{this seems to be a dataset face? define the unknown for this figure. The figure shows emotion - the generalisation property in what respect?}) 
(Fig. \ref{fig:qualitative_mead}, Row 3 \& 4, refer to the caption for evaluation details). Moreover, MEAD\footnote{\label{fn:mead1} https://github.com/uniBruce/Mead} and EVP\footnote{\label{fn:evp1} https://github.com/jixinya/EVP} train target-specific texture models.

In this work, we propose a generalized one-shot learning-based emotional talking face generation method. Unlike the previous video-based method EVP (Table \ref{tab:list_soa}), for emotion rendering, we need only a single image of the target person, along with speech and an emotion vector as input. We want to achieve speech-independent emotion control so that the same audio can be animated using different emotions. We use features from a pre-trained automatic speech recognition model DeepSpeech \cite{hannun2014deep} for disentangling emotion from speech content of audio. We first propose a graph neural network that encodes the desired emotion and speech content to render emotion and speech-induced motion on a geometry-aware graph representation of the facial landmarks.  Unlike previous landmark-based talking face methods 
\cite{chen2019hierarchical,zhou2020makelttalk,chen2020talking,ji2021audio,zhang2021flow}, we construct a graph representation of facial landmarks using \cite{delaunay1934bulletin} for capturing 
the spatial configuration of facial landmarks and their inter-dependencies during emotional speech. In the texture generation stage, we learn an emotion-guided optical flow map from the intermediate predicted landmarks to consider the facial structure and emotion-induced local deformations around the landmarks. Despite having high-quality, well-defined emotional speech videos, MEAD dataset has low variety in illumination, background, etc. We carefully design a two-branch texture generation network to disentangle the speech and emotion-induced motion from identity-related texture content. At inference time, we propose one-shot learning for adapting the texture generation model to the identity of the input target face. This helps in generalization while generating emotions for any arbitrary target face.

We demonstrate the generalization ability of our method by evaluating on different faces outside our training dataset MEAD (Fig.s \ref{fig:intro_img}, \ref{fig:suppfig2}, \ref{fig:suppfigback} and  \ref{fig:qualitative_crema}). %more examples in supplementary video\footnote{https://youtu.be/qqqRx-z5l9k}). 
To the best of our knowledge, this is the first work on emotional talking face generation that is generalized for any arbitrary face. Our contributions are summarized below:
\begin{itemize}
    \item We propose a pipeline for facial geometry-aware one-shot emotional talking face generation from audio with independent emotion control.
    \item We propose a graph convolutional network for inducing speech and emotion on graph-representation of facial landmarks to preserve facial structure and geometry for emotion rendering.
    \item We propose an optical flow-guided texture generation network that renders emotional talking face animation from a single image of any arbitrary target face in neutral emotion.
\end{itemize}

\begin{table}[t]
\centering
    \noindent\resizebox{\columnwidth}{!}{
\begin{tabular}{|l|c|c|c|}
\hline 
 \textbf{Audio-driven}  & \textbf{Input} & \textbf{Arbitrary} & \textbf{Emotion} \\ 
 \textbf{Talking Face Methods} & \textbf{Image/Video} & \textbf{face} & \textbf{generation} \\ 

 \hline

 \hline
 \cite{dasspeech} & Image & \checkmark & \xmark \\ 
 \hline
 
 MakeItTalk \cite{zhou2020makelttalk} & Image & \checkmark & \xmark  \\
  \hline
\cite{zhang2021flow} & Image & \checkmark & \xmark  \\
  \hline
\cite{wangaudio2head}& Image & \checkmark & \xmark  \\
  \hline
 \cite{zhou2021pose} & Image  & \checkmark & \xmark  \\
 %& source video & & \\
  \hline
\cite{thies2019neural} &  Video & \checkmark &  \xmark  \\
 \hline
 
\cite{song2020everybody} &  Video & \checkmark & \xmark  \\
\hline
 
Wav2Lip \cite{prajwal2020lip} &  Video & \checkmark & \xmark  \\
\hline
\cite{wen2020photorealistic} &  Video & \checkmark & \xmark  \\
\hline

\hline
\cite{vougioukas2019realistic}* & Image & \xmark &  \checkmark \\ %
 \hline
 \cite{chen2020talking}* & Image  & \xmark & \checkmark  \\
 %& Pose init video & &\\
 \hline
 \cite{eskimez2020speech} & Image & \xmark & \checkmark \\ %
 \hline
 MEAD, \cite{wang2020mead} & Image & \xmark & \checkmark \\
\hline
EVP, \cite{ji2021audio} & Video & \xmark & \checkmark  \\
\hline 
\hline
\textbf{Ours} & \textbf{Image} & \boldcheckmark & \boldcheckmark \\
\hline 
\end{tabular} 
}
\caption{\small {Recent talking face generation methods. 
%The methods which generalize well to unknown faces do not render emotion, while 
The emotional talking faces cannot generalize to arbitrary faces. (*) Emotion is not learned explicitly in these methods, derived implicitly from audio.}}
\label{tab:list_soa}
\end{table}

\section{Related Work}

\textbf{Emotional Talking Face Generation:}
Recent methods in audio-driven talking face generation are listed in (Table \ref{tab:list_soa}). 
Video-based methods that generate only the mouth in a driving video of target \cite{thies2019neural,song2020everybody,prajwal2020lip,wen2020photorealistic} are capable of generating photo-realistic facial animation. However, since the facial texture (except the mouth) is copied from the input video frames, facial expressions and emotions in the upper part of the face cannot be manipulated using these methods. %Moreover, it is infeasible for practical applications to obtain a sufficiently long driving video of a subject in different emotions. 
Our method uses a single image of the target for generating emotional talking faces without the need for a driving video.

%Few speech-driven animation methods have explicitly tried to render emotion due to the limitation of availability of generalized large-scale emotion annotated audio-visual datasets. 
Some earlier methods \cite{vougioukas2019realistic,chen2020talking} render emotional talking face videos that learn the emotion implicitly from the audio. In contrast, we aim for an explicit control for generating consistent emotions in the talking face.
%directly from emotional speech, without explicit modeling of speech. The emotion information is directly encoded in the audio feature representation, hence explicit control of emotion is not possible. 
Some recent methods MEAD, EVP, \cite{eskimez2020speech} have proposed methods with external control on emotion in the talking face. 
%A recent method learns 
EVP learns a disentangled emotion latent feature representation from speech input and tries to generate varying emotions by interpolating the emotion latent space. However, the latent emotion representation in EVP depends on the accuracy of the audio-emotion disentanglement; hence it is difficult to achieve completely independent control of emotion from speech. In contrast to the previous methods MEAD, EVP, our method manipulates emotions in the entire face using an emotion control input that is fully independent of the audio.

\textbf{Generalized Arbitrary-Subject Talking Face:}
%Earlier methods on audio-driven talking face generation generated target-specific facial animation \cite{suwajanakorn2017synthesizing} using a large amount of target-specific training data. 
Talking face generation methods (Table \ref{tab:list_soa}) that can generalize to arbitrary faces are trained on large-scale audio-visual datasets such as Voxceleb \cite{chung2018voxceleb2} having a wide diversity of faces, illumination and background. However these methods cannot render animation in different emotions. Existing emotional talking face generation methods trained on emotional audio-visual datasets CREMA-D \cite{cao2014crema} and MEAD \cite{wang2020mead} have limited scope of generalization owing to lower diversity of these datasets.
%, owing to the limitation of emotion-annotated audio-visual datasets. 
Previous methods \cite{vougioukas2019realistic,chen2020talking,eskimez2020speech} which are trained on CREMA-D lack generalization to faces outside CREMA-D. 
%Two recent methods 
Recently, MEAD and EVP have used a high quality emotional audio-visual dataset MEAD for training. However, they have trained target subject-specific texture generation models  \footref{fn:mead1} \footref{fn:evp1}; hence they cannot generalize to arbitrary identities. On the other hand, our method is capable of generalization to any unknown target subject.

\begin{figure*}[t]
    \centering
    \includegraphics[width=\textwidth]{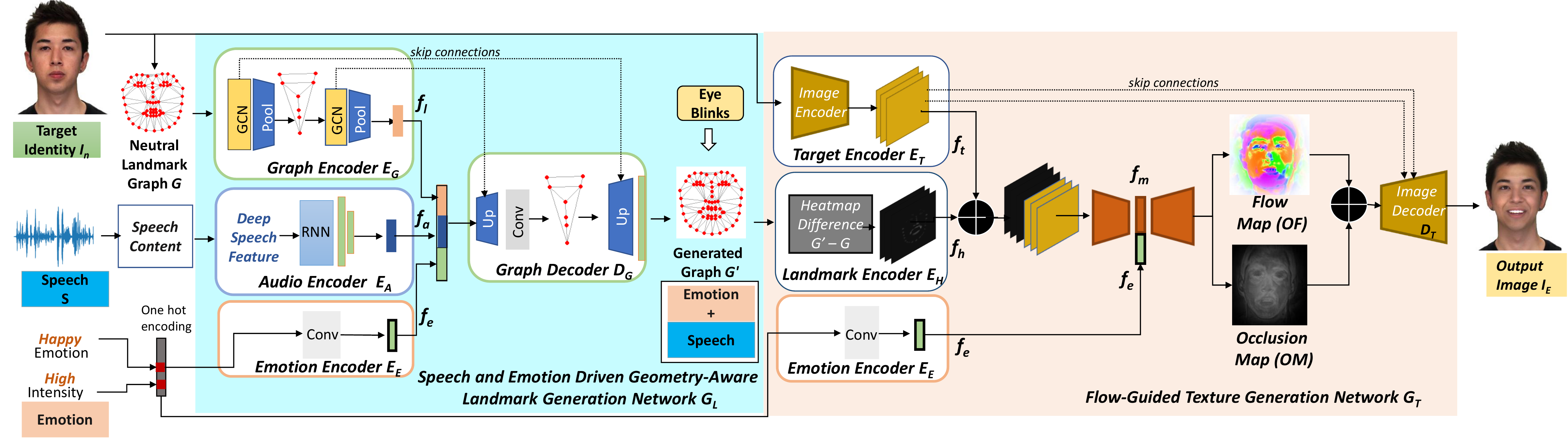}
    \caption{\small{Our proposed method for arbitrary-face emotional talking face generation. The Geometry-Aware Landmark Generation Network ${G}_{L}$, encodes speech content of input speech $S$, neutral face landmark graph $\mathcal{G}$, target emotion $e$ (along with emotion intensity), and reconstructs landmark graph $\mathcal{G'}$ containing speech and emotion. For realism spontaneous, eye blinks are added to the landmarks in $\mathcal{G'}$. In the Texture Generation stage, the heatmap difference of the target identity's facial landmarks, encoded identity face, and encoded target emotion are used to generate emotion-induced optical flow and occlusion map, which are subsequently decoded to generate the speech and emotion-induced facial texture image of the target identity.}}
    \label{fig:blockdiagram}
\end{figure*}

\section{Methodology}

Fig. \ref{fig:blockdiagram} shows the detailed architecture of our network for generating emotion-controllable talking faces. For a given speech ($S$), an emotion input, and a single image of the target subject in neutral emotion ($I_n$), our method generates an animated face delivering the speech with desired emotion and intensity.

\subsection{Speech and Emotion Driven Landmark Generation}
%Figure \ref{fig:blockdiagram} shows the detailed architecture of 
We propose facial geometry-aware speech and emotion generation (${G}_{L}$, Fig. \ref{fig:blockdiagram}) on facial landmarks using a graph neural network.  
\\
\textbf{Audio Encoder, $E_A$} is a recurrent neural network which creates an emotion-invariant speech embedding feature $\mathbf{f_a} \in \mathbb{R}^d$ ($d=128$) from speech audio input $S$.
%DeepSpeech \cite{hannun2014deep} encodes the speech content from audio, which is independent of the emotional content of the audio.  
For each audio window of size $W$ corresponding to a video frame, features $\mathcal{A} = \{a_t \in \mathbb{R}^{W \times 29}\}$ are extracted from the output layer of a pre-trained DeepSpeech network (before applying Softmax). The output layer of Deepspeech represents log probabilities of $29$ characters; hence the features are emotion-independent. 
\\
%The Audio Encoder encodes $a_t$ to
%encodes input Deepspeech features $a_t \in \mathbb{R}^{W \times 29}\ $ to 
%$E_a(a_t) = F_a \in \mathbb{R}^d$, where $F_a$ is a $d$-dimensional feature to be used further for feature concatenation before graph reconstruction.
\textbf{Emotion Encoder}, $E_E$ encodes an emotion vector ($e$,$i$). $e$ denotes six types of emotions i.e. happy, angry, sad, surprise, fear and disgust, at two types of intensity levels $i$ (high or low) into a fixed feature representation $\mathbf{f_e} \in R^d$ ($d=128$).% for concatenation with audio and graph features. We encode the emotion and intensity input using one-hot encoding and the concatenated emotion vector $\mathbf{e}$ is encoded to a $d$-dimensional emotion feature $E_e(e,i) = F_e \in R^d$.
\\
\textbf{Graph Encoder, ${E}_{G}$} is a graph convolutional network that encodes the geometry of an ordered graph $\mathcal{G}= (\mathcal{V}, \mathcal{E}, A)$, where $\mathcal{V} = \{v_{i}\}$ denotes the set of $L=68$ facial landmark vertices, $\mathcal{E} = \{e_{ij}\}$ is the set of edges, computed using delaunay triangulation \cite{delaunay1934bulletin} on facial landmarks, $A$ is the adjacency matrix of $\mathcal{G}$. $\mathbf{X} = [X_{ij}](X_{ij} \in \mathbb{R}^{2})$ is a matrix of vertex feature vectors, i.e, coordinates of the $L=68$ facial landmarks of a neutral image (face in neutral emotion and with closed lips). We apply spectral graph convolution \cite{kipf2016semi} with the following modified propagation rule including learnable edge weights \cite{yan2018spatial} :
\begin{equation}
\label{eqn:graphcnn}
    f_{k+1} = \sigma (\Tilde{D}^{-\frac{1}{2}}\boldsymbol{\omega}(A+I) \Tilde{D}^{-\frac{1}{2}}f_{k}W_{k} ),
\end{equation}

where %$\hat{A} = \mathcal{\boldsymbol{\omega}}\mathcal{A} $,
 $I$ represents the identity matrix,  $\Tilde{D}^{ii}= \sum_{j}(A^{ij}+I^{ij} )$, $\boldsymbol{\omega}=\{\omega^{ij}\}$ are learnable edge weights for determining the contribution of each edge in $\mathcal{G}$,  $f_{k}$ is the output of the $k$th layer, ($f_{0} = \mathbf{X}$), $W_{k}$ is a trainable weight matrix of the $k$th layer, $\sigma(\cdot)$ is the activation function. Since edges between landmark vertices of semantically connected regions of the face are more significant than the edges connecting two different facial regions, the learnable edge weight $\boldsymbol{\omega}$ signifies the contribution of the vertex's feature to its neighboring vertices. %Experimentally (Section \ref{sec:ablation}) we find that using $\boldsymbol{\omega}$ improves accuracy of generating facial expressions (refer Ablation Study).
Unlike lip movements, emotion has an effect over the entire face and not only a specific region. Inspired by \cite{cai2019exploiting} we apply a hierarchical “local-to-global” scheme for graph convolution to capture facial deformations. %The face landmark graph structure is first divided into $K$ subsets of vertices, each representing a facial region, e.g., eye, nose, etc. Graph max-pooling is performed on each vertex set $\{V_{i}|i=1,2 \cdots K\}$ to generate features of a smaller graph $G^{1}$ with $K$ vertices. 
Graph pooling operation helps to aggregate feature level information in different facial regions, which helps local deformations caused by facial expressions. The face landmark graph structure is first divided into $K$ subsets of vertices, each representing a facial region, e.g., eye, nose, etc. 
%Graph pooling is performed on each vertex set $\{V_{i}|i=1,2 \cdots K\}$ to generate features of a smaller graph $G^{1}$ with $K$ vertices. Further graph convolution and pooling are done to create a $d$ dimensional feature $F_l \in R^d $ representing a graph $G^{2}$ consisting of a single vertex representing the entire face. 
Hierarchical graph convolution (GCN) and pooling is done (as shown in Fig. \ref{fig:blockdiagram}) to generate feature $\mathbf{f_l} \in R^d $ ($d=128$) representing the entire graph.
\\
\textbf{Graph Decoder}, ${D}_G$ reconstructs the output landmark graph $\mathcal{G'}= (\mathcal{V'}, \mathcal{E}, A)$ from the concatenation of the feature vectors $\mathbf{f_a},\mathbf{f_l},\mathbf{f_e}$. It learns the mapping $f:(\mathbf{f_a},\mathbf{f_l},\mathbf{f_e}) \rightarrow \mathbf{X'}$, where $\mathbf{X'} = \mathbf{X} +\boldsymbol{\delta}$ represents the vertex positions of the reconstructed facial landmarks with generated displacements $\boldsymbol{\delta}$ induced by speech and emotion. $\mathbf{\hat{X}}$ are the ground landmarks.
%Graph upsampling is performed in $D_G$ to reconstruct the final graph. Graph convolution intermediate features in $E_G$ are added via skip connections during upsampling in ${D}_G$ for preserving facial geometry-related information during graph reconstruction.
\\
The losses for training ${G}_{L}$ are as follows:\\
\textit{Landmark vertex distance loss}: 
%The distance between predicted and ground-truth landmarks are minimized using the following loss function:
% \begin{equation}
% \mathcal L_{ver} = \sum_{l=1}^{L} ||\hat{X}_{l} - (X_{l} + \delta_{l})||^2_2
% \end{equation}
\begin{equation}
\mathcal L_{ver} =  ||\mathbf{\hat{X}} - (\mathbf{X} + \mathbf{\delta})||^2_2.
\end{equation}

\textit{Adversarial loss}:
 
A graph discriminator ${D}_{L}$ evaluates the realism of the facial expression in a generated graph $\mathcal{G'}$. ${G}_{L}$ and ${D}_{L}$ are trained using the LSGAN loss function \cite{mao2017least}:
%\begin{equation}
\begin{align}
\mathcal L_{gan}(D_{L}) &= \big( \mathbb{E}[(D_{L}(\mathcal{\hat{G}},e)-1)^2] +  \mathbb{E}[D_{L}(\mathcal{G'},e)^2] \big) / 2 \nonumber \\ 
\mathcal L_{gan}(G_{L}) &=\mathbb{E}[(D_{L}(\mathcal{G'},e)-1)^2]/2,
\end{align}
%\end{equation}
where $\mathcal{G'}$ is the generated graph and $\mathcal{\hat{G}}$ is the ground truth graph.
The combined loss function for training the landmark generation networks are:
\begin{equation}
\label{eqn:loss_landmark}
\mathcal L_{lm} = \lambda_{ver} L_{ver} +  \lambda_{gan}L_{gan},
\end{equation}

where the loss hyperparameters $\lambda_{ver}=1$ and $\lambda_{gan}=0.5$ are experimentally set using validation data.

\subsection{Texture Generation}

Fig. \ref{fig:blockdiagram} shows our proposed Texture Generation network $G_T$ that generates an emotional talking face from a single image $I_n$ of the target identity subject in neutral expression and predicted landmarks $\mathcal{G'}$ from $G_L$. For realism, spontaneous eye blink displacements \cite{dasspeech} are added to the landmark vertices of $\mathcal{G'}$ before texture generation. %(Refer \textbf{\textit{Technical Appendix (TA) Sec. 1.1).}}
%Next we generate the facial texture with motions due to speech and emotion.  
%The texture generator module takes the predicted landmarks from the previous stage, a target emotion, and emotion intensity as input. 
\\
\textbf{Image Encoder, $E_T$} encodes the target identity image $I_n$ into identity feature $f_t$, that is used for predicting the optical flow and occlusion map in the subsequent stage. The emotion feature $f_e$ is generated in a similar manner as presented in the landmark generation network $G_L$.\\
\textbf{Heatmap Difference:} A heatmap is generated by creating a Gaussian distribution centered at each of the vertices of the landmark graph. The heatmap representation captures the structural information of the face in the image space and the local deformations around the landmark vertices. The difference $f_h$ between heatmaps of input graph $\mathcal{G}$ and generated graph $\mathcal{G}'$ is computed to model the motion of facial landmarks. \\ 
\textbf{Optical Flow and Occlusion Map Prediction:} Optical flow ($OF$) captures the local deformations over different regions of the face due to speech and emotion induced motions. Whereas, occlusion map ($OM$) denotes the regions which need to be newly generated (e.g., inside the mouth region for happy emotion) in the final texture. $OF$ and $OM$ are learned in an unsupervised manner (Eqn. \ref{eqn:FA}) and no ground-truth optical flow or occlusion map are used for supervision. At an intermediate stage the network generates $OF$ and $OM$ from heatmap difference, target identity image conditioned on emotion condition. The heatmap difference ($f_h$) and the encoded target identity image feature ($f_t$) are concatenated channel-wise and passed through an encoder network to produce $f_m$. Further, to influence the facial motion by the necessary emotion, the encoded emotion feature $f_e$ is concatenated channel-wise with $f_m$ and decoded to produce the dense flow map ($OF$) and occlusion map ($OM$). Flow-guided texture generation from heatmap differences of facial landmarks helps to learn the relationship between the face geometry and emotion-related deformations within the face. \\
\textbf{Final Animation Generation:} The concatenated occlusion map and optical flow maps are given as input to the image decoder $D_T$, which produces the final output image ($I_E$) containing speech and emotion.
\begin{equation}
I_E = D_T(OF \oplus OM,f_t).
\label{eqn:FA}
\end{equation}
%The encoder layers ($E_I$) features are used to influence the decoder layer weights using Adaptive Instance Normalization (AdaIn) \cite{huang2017arbitrary}. 
Skip connections are added between the layers of target identity encoder ($E_T$) and the decoder $D_T$. 
The losses used for training the network are as follows:\\
\textit{Reconstruction loss} between predicted $I_E$ and GT image $\hat{I}$:
\begin{equation}
    \mathcal L_{rec} = |I_E - \hat{I}|.
\end{equation}
\textit{Perceptual loss} between VGG16 features of $I_E$ and $\hat{I}$:
\begin{equation}
    \mathcal L_{per} = |VGG16(I_e) - VGG16(\hat{I})|.
\end{equation}
\textit{Adversarial loss} with a frame discriminator $D$:
\begin{equation}
    \mathcal L_{adv} = \min\limits_{G} \max\limits_{D} \mathbb{E}_{\hat{I}} [log(D(\hat{I}))] + \mathbb{E}_{I_E}[log (1-D(I_E))]. 
\end{equation}
The total loss function for training $G_T$:
\begin{equation}
    \mathcal L_{img} = \lambda_{rec} L_{rec}+\lambda_{per} L_{per}+\lambda_{adv} L_{adv},
\end{equation}
where the loss hyperparameters $\lambda_{rec}$, $\lambda_{per}$, $\lambda_{adv}$ are experimentally set to 1, 10, and 1 respectively.

\begin{table*}[t]
\centering
    \noindent\resizebox{0.9\textwidth}{!}{
\begin{tabular}{|c|c|cccr|cccc|c|c|c|}
\hline 
Dataset & Method & \multicolumn{4}{c|}{\textbf{Texture Quality}} & \multicolumn{4}{c|}{\textbf{Landmark quality}} & \textbf{Emotion accuracy} & \textbf{Identity} & \textbf{Lip Sync} \\
& & PSNR & SSIM & CPBD & FID & M-LD & M-LVD & F-LD & F-LVD & $Emo_{Acc}$ & CSIM & $Sync_{conf}$\\ 
\hline

\multirow{2}{*}{MEAD} & MEAD \cite{wang2020mead}  & 28.61 & 0.68 & 0.29 & 22.52 & 2.52 & 2.28 & 3.16 & 2.01 & 76.00 & \textbf{0.86} & 1.83 \\

&  EVP \cite{ji2021audio} & 29.53 & 0.71 & 0.35 & \textbf{7.99} & 2.45 & 1.78 & 3.01 & 1.56 & 83.58 & 0.67 & 1.21\\
&  Ours & \textbf{30.06} & \textbf{0.77} & \textbf{0.37} & 35.41 & \textbf{2.18} & \textbf{0.77} & \textbf{1.24} & \textbf{0.50} & \textbf{85.48} & 0.79 & \textbf{3.05}\\ %\textbf{3.541(2.26})
\hline
% \textbf{Pose-controlled} \cite{zhou2021pose} &  &  &  &  &  &  &  &  & &&\\
% \hline 
% Flow-guided \cite{zhang2021flow} &  &  &  &  &  &  &  &  & &&\\
% \hline

\multirow{2}{*}{CREMA-D} &\cite{vougioukas2019realistic}  & 23.57 & 0.70 & 0.22 & 71.12 & 2.90 & \textbf{0.42} & 2.80 & \textbf{0.34} & 55.26 & 0.51 & 1.12\\

& \cite{eskimez2020speech}  & 30.91 & 0.85 & 0.39 & 218.59 & 6.14 & 0.49 & 5.89 & 0.40 & 65.67 & \textbf{0.75} & \textbf{4.38} \\
%& \textbf{Chen et al.} \cite{chen2020talking}  & & &  &  &  &  &  &  & &  & \\
&  Ours & \textbf{31.07} & \textbf{0.90} & \textbf{0.46} & \textbf{68.45} & \textbf{2.41} & 0.69 & \textbf{1.35} & 0.46 & \textbf{75.02} & \textbf{0.75} & 3.53
\\
\hline
\end{tabular}
}

\caption{\small{Quantitative comparison of our method with SOTA emotional talking face generation methods.  \protect\cite{eskimez2020speech,vougioukas2019realistic} have trained their method on CREMA-D dataset, while MEAD, EVP have trained on MEAD dataset.  Our model is trained only on MEAD and evaluated on both MEAD and CREMA-D.}}
\label{tab:quantitative}
\end{table*}

\section{Experiments and Training Details}
\textbf{Datasets}
We use 3 emotional audio-visual datasets MEAD \cite{wang2020mead}, CREMA-D \cite{cao2014crema}, and RAVDESS \cite{livingstone2018ryerson} for our experiments.   
We have selected 24 subjects of diverse ethnicity from MEAD for the training of our proposed pipeline, and our method is evaluated on test splits of MEAD, CREMA-D, RAVDESS and also arbitrary unknown faces and speech.

\subsection{Implementation Details:}

%We use \cite{guo2020towards,3ddfa_cleardusk} for $3D$ facial landmark extraction from the ground-truth videos at $30 fps$. 
The Landmark Generation Network $G_L$ and Texture Generation Network $G_T$ are trained independently. The architectures of $G_L$ and $G_T$ are shown in Fig. \ref{fig:blockdiagram}. % are described in the \textcolor{red}{\textbf{\textit{TA (Sec. 2.3)}}}. 
%We register the $3D$ landmarks to a neutral frontal canonical landmark (mean shape over the dataset) for training the landmark generation network. 
For training $G_L$ and $G_T$, the ground-truth landmarks are extracted (at $30 fps$) using a combination of 3D landmarks from \cite{guo2020towards} and face parsing \cite{yu2018bisenet} for accurate mouth shapes.  %Detailed data pre-processing steps are given in the \textbf{\textit{TA (Sec. 2.2)}}.
$G_T$ uses ground-truth landmarks during training, and predicted landmarks from $G_L$ during inference. %At inference time, our complete pipeline takes as input a single image of target, speech, emotion type and intensity, and generates texture corresponding to the speech and emotion input.
We train both $G_L$ and $G_T$ using Pytorch on NVIDIA Quadro P5000 GPUs (16 GB) using Adam Optimizer, with a learning rate of $2e-4$. Training of $G_L$ takes  around  a  day  with  batch  size $256$ ($2$GB GPU usage), and the training of $G_T$ takes around $7$ days (batch size $4$ on $16$GB GPU). \\ %We train the audio to emotion-induced landmark generation module on frontal canonical face. For this purpose, we frontalize the person-specific facial landmarks (extracted from ground-truth videos) to add the displacements (generated from the facial motions due to speech and emotion) of person-specific landmarks with a static canonical frontal face (mean face over the dataset). \\
\textbf{One-shot learning:} MEAD dataset contains a limited variety in illumination and background, which limits generalization to arbitrary target faces. By fine-tuning our texture generation network $G_T$ on a single image of any unseen target face \textit{in neutral emotion}, we can generate emotional talking face generation for the target \textit{in different emotions}. 
In order to adapt to the identity of the unknown target neutral, we only update the image encoder ($E_T$) and decoder layers ($D_T$) weights using the single image in neutral emotion, while keeping the network weights for the rest of $G_T$ unchanged. This fine-tuning is done for upto $5$ iterations, and it takes around $3-4$ seconds. One-shot learning helps bridge the color and illumination gap between the training and testing samples and adapt the generated texture to the identity of the target face while keeping the speech and emotion-induced motion intact. %The significance of our one-shot learning for emotional talking face generalization is described in the \textbf{Appendix}. % \textcolor{red} {\textbf{\textit{TA (Sec. 2.5)}}}.

\begin{figure*}[t]
    \centering
    \begin{minipage}{.62\textwidth}
        
        \includegraphics[width=\linewidth]{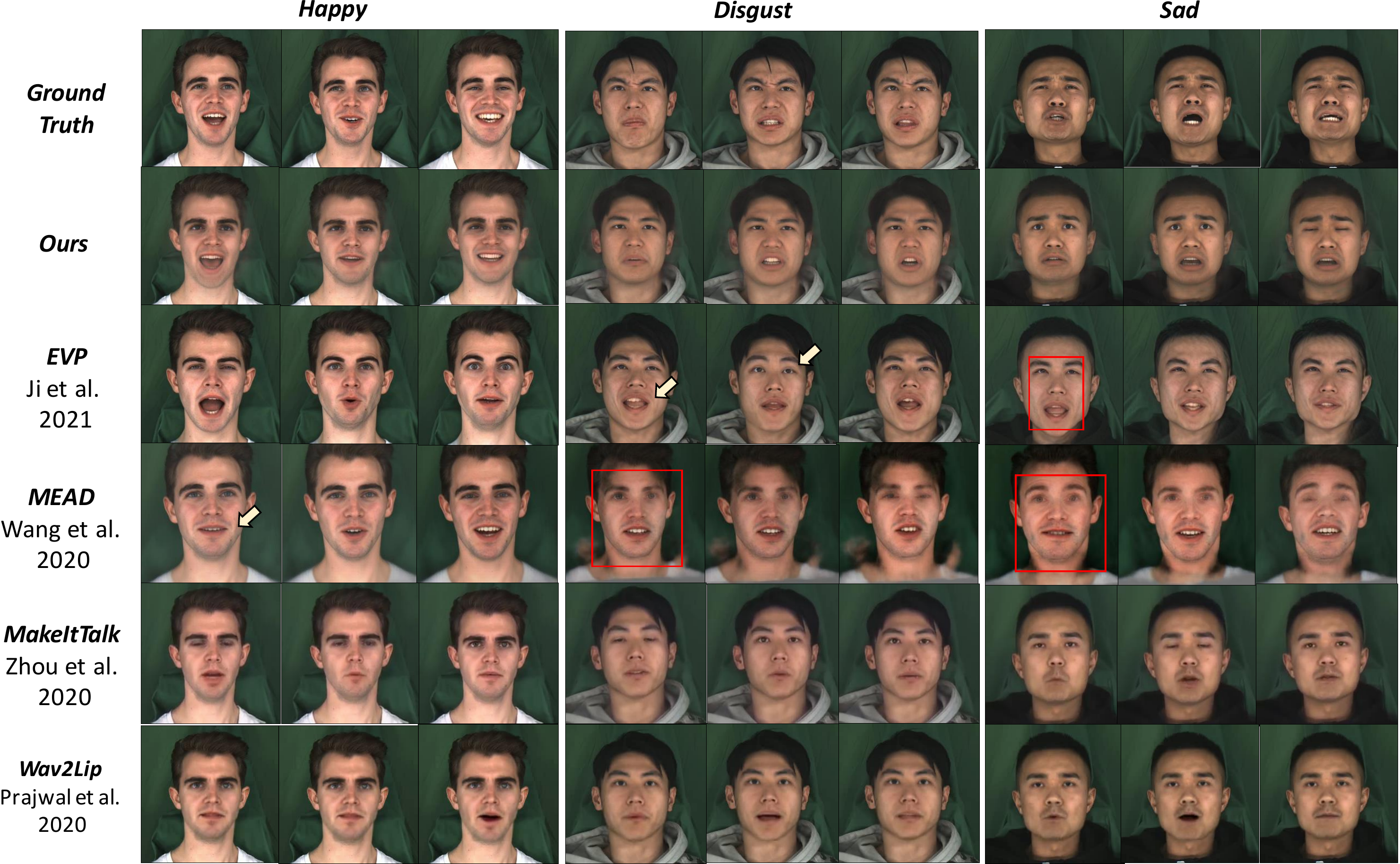}
        \caption{\small{Qualitative comparison of our method with SOTA on MEAD dataset. MakeItTalk and Wav2Lip do not render emotion. Since the publicly available pre-trained model for MEAD \protect \footnoteref{fn:mead} %\footref{fn:mead}
 is only trained for Subject 1 (left), their method is unable to generalize to other identities (in red box). Similarly for EVP, the publicly available target-specific pre-trained texture models  \protect \footnoteref{fn:evp} %\footref{fn:evp} 
are available only for Subjects 1,2 (left and middle). Hence their method fails to generalize to Subject 3 (right) as shown in red box (Subject 3 evaluated using a pre-trained model for Subject 2). The white arrow shows inconsistent emotions at the mouth and eyebrow regions.}}
        \label{fig:qualitative_mead}
        
    \end{minipage}%
    \hspace{3mm}
    \begin{minipage}{0.35\textwidth}
        \centering
       
         \begin{minipage}{\linewidth}
          \captionsetup{type=table} 
         \noindent\resizebox{\columnwidth}{!}{
        \begin{tabular}{|c|c|c|c|c|}
        \hline 
            Methods & M-LD & M-LVD & F-LD & F-LVD \\ 
            \hline
            Ours w/o Graph Encoder $E_a$  &5.54 & 0.54 & 2.75 &  0.43  \\ 
            \hline
            Ours w/o skip connections &5.54 & 0.54 & 2.75 & 0.43  \\ 
            \hline 
            Ours w/o edge weights $\omega$ & 2.45 & 0.83 & 1.39 & 0.52 \\ 
            \hline 
            Ours w/o $L_{gan}$ & 2.52 & 0.86 & 1.42 & 0.53  \\
            \hline 
            \textbf{Ours} & \textbf{2.18} & \textbf{0.77} & \textbf{1.24} & \textbf{0.5}  \\
            \hline 
        \end{tabular}
        }
        \caption{\small {Ablation study for Landmark Generation.}}
        \label{tab:ablation_landmark}
        \end{minipage}
        \hfill
        \vspace{2mm}
        \begin{minipage}{\linewidth}
         \captionsetup{type=table} 
        \noindent\resizebox{\columnwidth}{!}{
        \begin{tabular}{|c|c|c|c|}
            \hline 
            \textbf{Methods} & \textbf{PSNR} & \textbf{CSIM} & \textbf{Emotion Acc.} \\ 
            \hline 
            Ours w/o emotion feature & 29.83 & \textbf{0.885} & 45.00 \\ 
            \hline 
            Ours w/o emotional landmark & 29.85 & 0.861 & 59.61 \\ 
            \hline 
            Ours w/o one-shot learning & 29.89 & 0.767 & 84.00 \\
            \hline
            Ours & \textbf{30.06} & 0.789 & \textbf{85.48} \\ 
            \hline 
            % W/o Data Augmentation & • & • & • \\ 
            % \hline 
            \end{tabular} 
        }
        \caption{\small {Ablation study for Texture Generation.}}
        \label{tab:ablation_texture}
        \end{minipage}
        \hfill
        
        \begin{minipage}{\linewidth}
        \vspace{4mm}
            \centering
            \includegraphics[width=0.9\columnwidth]{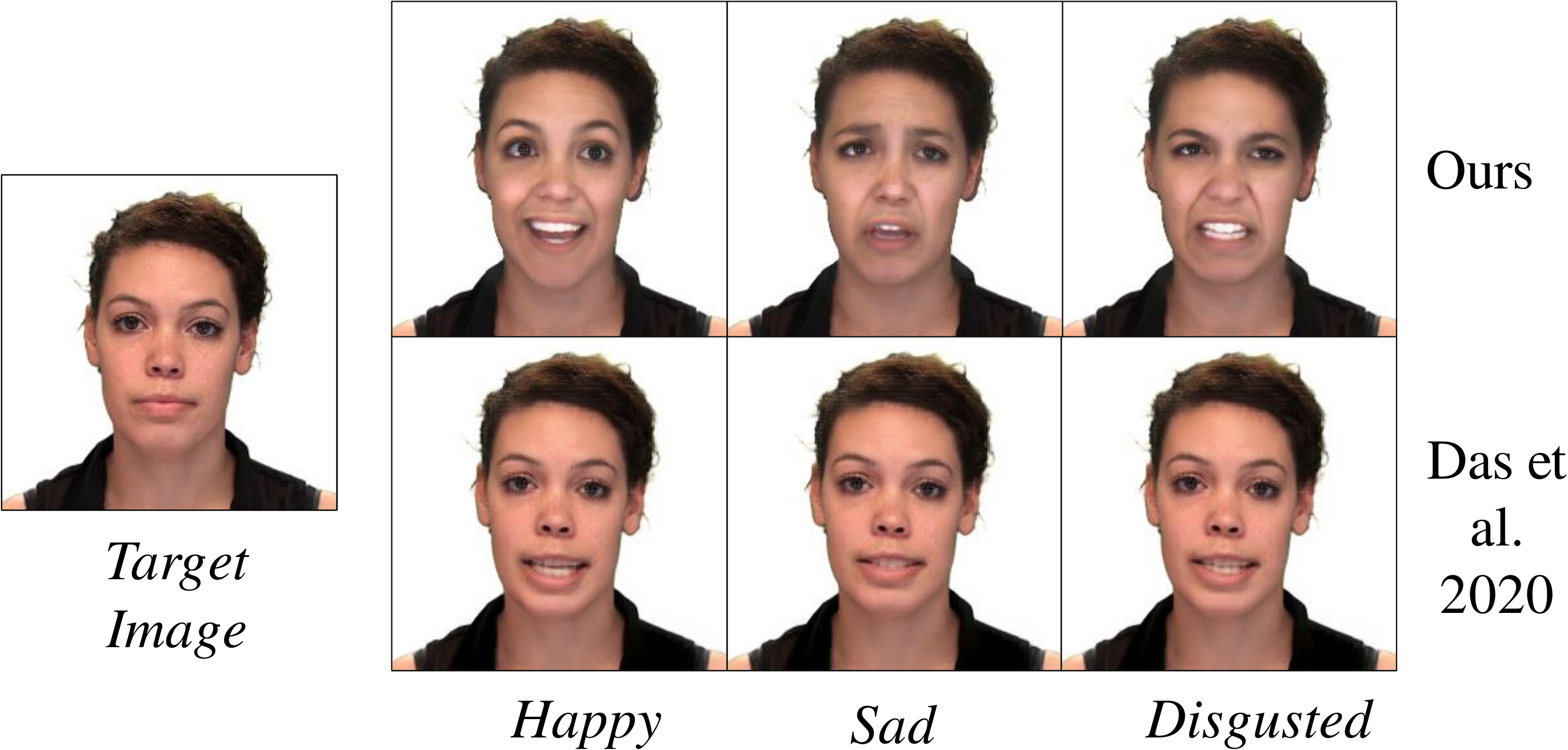}
            \caption{\small{Comparison of one-shot learning with \protect\cite{dasspeech} on a subject from RAVDESS. Our model is trained on MEAD and generates emotions using one-shot learning on a target identity (in neutral emotion) from RAVDESS.}} %in \cite{vougioukas2019realistic,eskimez2020speech} can also be attributed to low variety of short and simple sentences in CREMA-D dataset on which these methods are trained and evaluated. }}
            \label{fig:suppfig2}

        \end{minipage}
    \end{minipage}
\end{figure*}

\begin{figure*}[tb]
  \centering
  \includegraphics[width=\textwidth,trim={0cm 0cm 0 0cm},clip]{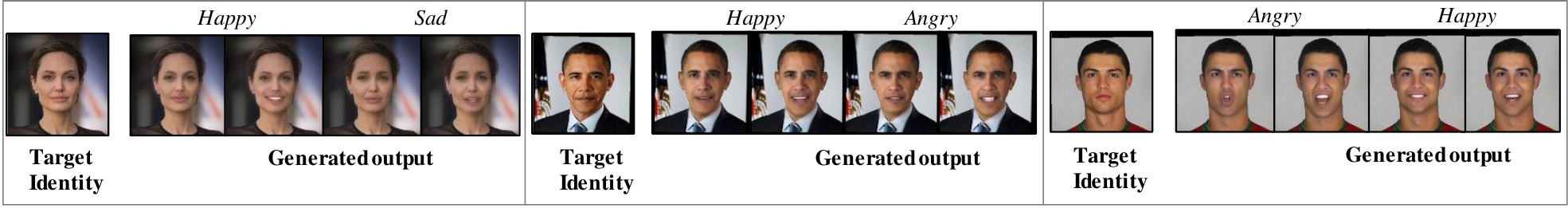}
  \caption{\small{Results in different emotions on arbitrary target faces with different backgrounds. Our texture generation network is trained on MEAD, which has a fixed background. To handle variable backgrounds, we replace the background of the input image of the target identity with the fixed background of MEAD. The background of the generated texture is substituted with the original background of the input image to produce the final output.  %(b) Results of one-shot fine-tuning of EVP \protect\cite{ji2021audio} with single neutral face image from MEAD.
  }}
  \label{fig:suppfigback}
  \end{figure*}

\subsection{Quantitative Results:} We evaluate our animation results against the state-of-the-art (SOTA) emotional talking face generation methods for assessing all the essential attributes of a talking face, i.e., texture quality, lip sync, identity preservation, landmark accuracy, the accuracy of emotion generation, etc. We present the quantitative results in Table ~\ref{tab:quantitative}. The emotional talking face SOTA methods MEAD, EVP, \cite{eskimez2020speech,vougioukas2019realistic} are dataset-specific and do not generalize well for arbitrary identities outside the training dataset. For a fair comparison, the evaluation metrics of SOTA methods have been reported for the respective dataset on which they were trained. %Please refer to the \textcolor{red}{\textbf{\textit{TA (Sec. 2.4)}}} for detailed explanation on why cross-dataset evaluation is not done for SOTA.  
However, the performance of our method is not restricted to the training dataset. Our method is trained only on MEAD dataset, but evaluated on both MEAD and CREMA-D.  
 The metrics used for the quantitative analysis are as follows:\\
\textbf{Texture quality}: We have used PSNR, SSIM \cite{wang2004image}, CPBD \cite{narvekar2009no}, and FID \cite{heusel2017gans} for quantifying the texture quality of the synthesized image. Our method outperforms the SOTA methods in most of the texture quality metrics. EVP outperforms all the methods in FID because they train person-specific texture models. \\
\textbf{Landmark quality}: We use Landmark Distance (LD) and Landmark Velocity Difference (LVD) \cite{ji2021audio} to quantify the accuracy of lip displacements (M-LD and M-LVD) and facial expressions (F-LD and F-LVD) with respect to the GT. On the CREMA-D dataset, although our velocity error metrics are slightly higher than SOTA methods, our landmark distance error metrics are much lower than the SOTA, indicating more accurate animation. \\
\textbf{Identity preservation}: We compute CSIM(cosine similarity) distance between ArcFace features \cite{deng2019arcface} of the predicted frame and the input identity face of the target. Our method outperforms MEAD. EVP outperforms our method in CSIM as they train texture models specific to each target identity. On the other hand, we use a single generalized texture model for all identities. 
% For MEAD and EVP, the reported values are calculated only for the subjects, for which pre-trained models are publicly available  \footnote{\label{fn:mead}https://github.com/uniBruce/Mead} \footnote{\label{fn:evp}https://github.com/jixinya/EVP} (Subject 1 in Figure \ref{fig:qualitative_mead} for MEAD and Subject 1 and 2 for EVP). In contrast, we evaluate on the entire test set of MEAD. 
Our one-shot learning helps to generalize on different subjects using only a single image of the target identity at inference time. Whereas EVP \footnote{\label{fn:evp}https://github.com/jixinya/EVP} and MEAD\footnote{\label{fn:mead}https://github.com/uniBruce/Mead} require sample images of the target in different emotions for training their target-specific models. \\
%Even with one-shot learning, our method can preserve the target \textbf{Lip sync.}: We use Syncnet \cite{Chung16a} to estimate the audio-visual synchronization accuracy in the synthesized videos. Our method can achieve better lip-sync compared to these methods.os. Our method can achieve better lip sync. compared to these methods.\\
\textbf{Emotion Accuracy}: We have used the emotion classifier network in EVP \cite{ji2021audio} for quantifying the accuracy of generated emotions in the final animation. %We have first trained this classifier using ground-truth frames of MEAD and CREMA-D datasets and tested on synthesized frames. 
On both the MEAD and CREMA-D datasets, we achieve better emotion classification accuracy than that of the existing methods. 
\\
\textbf{Audio-Visual Synchronization}: We use SyncNet \cite{Chung16a} to estimate the audio-visual synchronization accuracy in the synthesized videos. Our method achieves better lip sync than both EVP and MEAD on MEAD dataset, and performs better than \cite{vougioukas2019realistic} on CREMA-D. \cite{vougioukas2019realistic,eskimez2020speech} are trained on CREMA-D, whereas our method is trained on MEAD and evaluated on CREMA-D.

\begin{figure}[tb]
    \centering
    \includegraphics[width=0.95\columnwidth]{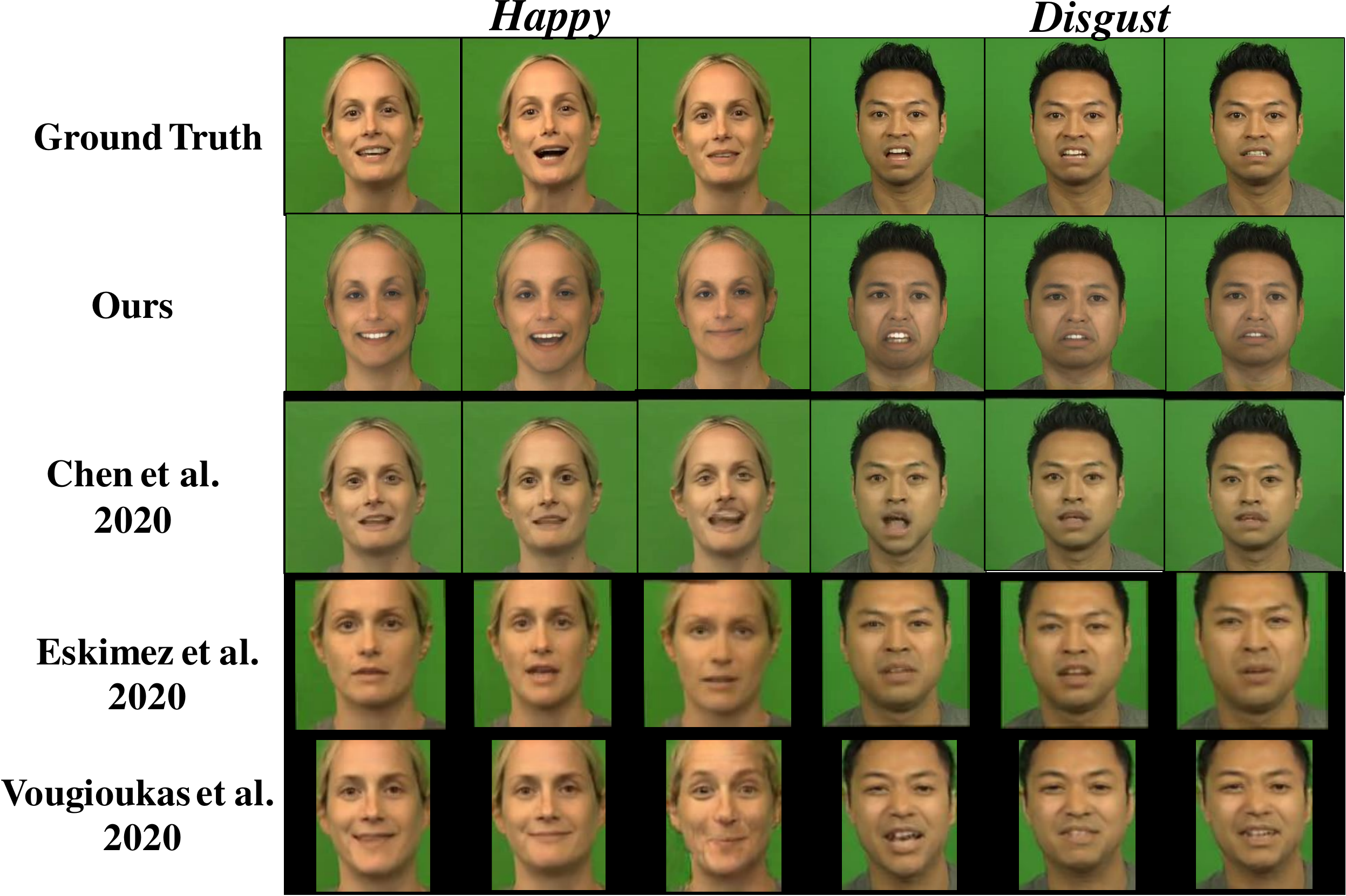}%{LaTeX/crema-crop.pdf}
    \caption{\small{Qualitative comparison on CREMA-D dataset. All the above SOTA methods (except \protect\cite{chen2020talking}) are trained on CREMA-D. \protect\cite{eskimez2020speech} is unable to generate significant emotion. \protect\cite{chen2020talking} produces distorted textures.}}
    \label{fig:qualitative_crema}
\end{figure}

\subsection{Qualitative Evaluation:}
\label{sec:qual}
Fig. \ref{fig:qualitative_mead} shows our final animation results on MEAD dataset compared to the recent SOTA methods MEAD, EVP, MakeitTalk\cite{zhou2020makelttalk} and Wav2Lip\cite{prajwal2020lip}. MEAD and EVP  are the most relevant works since they render emotion. We have evaluated MEAD using their publicly available pre-trained model \footref{fn:mead}, which is specific to subject 1 (First three columns) and fails to generalize for other subjects (column 4 to 9). %In EVP, the 3D geometry aware texture generation network EVP requires fine-tuning on emotion-induced samples of the target subject. 
EVP fails to preserve the identity of the target subject 3 (columns 7 to 9) without fine-tuning \footref{fn:evp}. Also, this method uses a latent feature learned from audio for emotion control, which makes the expressions inconsistent (happy emotion can be perceived as surprised or angry for subject 1, columns 1 to 3). Our method can produce better emotion and preserve identity even with one-shot learning using only a single neutral face image of the target person. Fig. \ref{fig:qualitative_crema} shows the comparative results on CREMA-D. Our method can produce realistic emotions on identities from other datasets, such as RAVDESS (Fig. \ref{fig:intro_img} upper face and Fig. \ref{fig:suppfig2}) as well as arbitrary faces (Fig. \ref{fig:intro_img} lower face and Fig. \ref{fig:suppfigback}). %More results are available in the \textcolor{red}{\textbf{\textit{supplementary video}}}.

 %\label{sec:supp-oneshot}

\textit{Efficacy of one-shot learning:} Fig. \ref{fig:suppfig2} shows a qualitative comparison with a recent talking face generation method \cite{dasspeech} that uses few-shot learning to adapt to arbitrary faces. For evaluation of \cite{dasspeech} under one-shot learning, we fine-tune their meta-learned texture model using a single image of a target face (in neutral emotion) from RAVDESS dataset. As shown in Fig. \ref{fig:suppfig2}, similar to \cite{dasspeech} our method can adapt to the identity of the target face. However unlike \cite{dasspeech}, using a single neutral emotion image for fine-tuning, our method can generate different emotions.  %

\subsection{Ablation Study:}
\label{sec:ablation}
\begin{figure}[tb]
    \centering
    \includegraphics[width=\columnwidth]{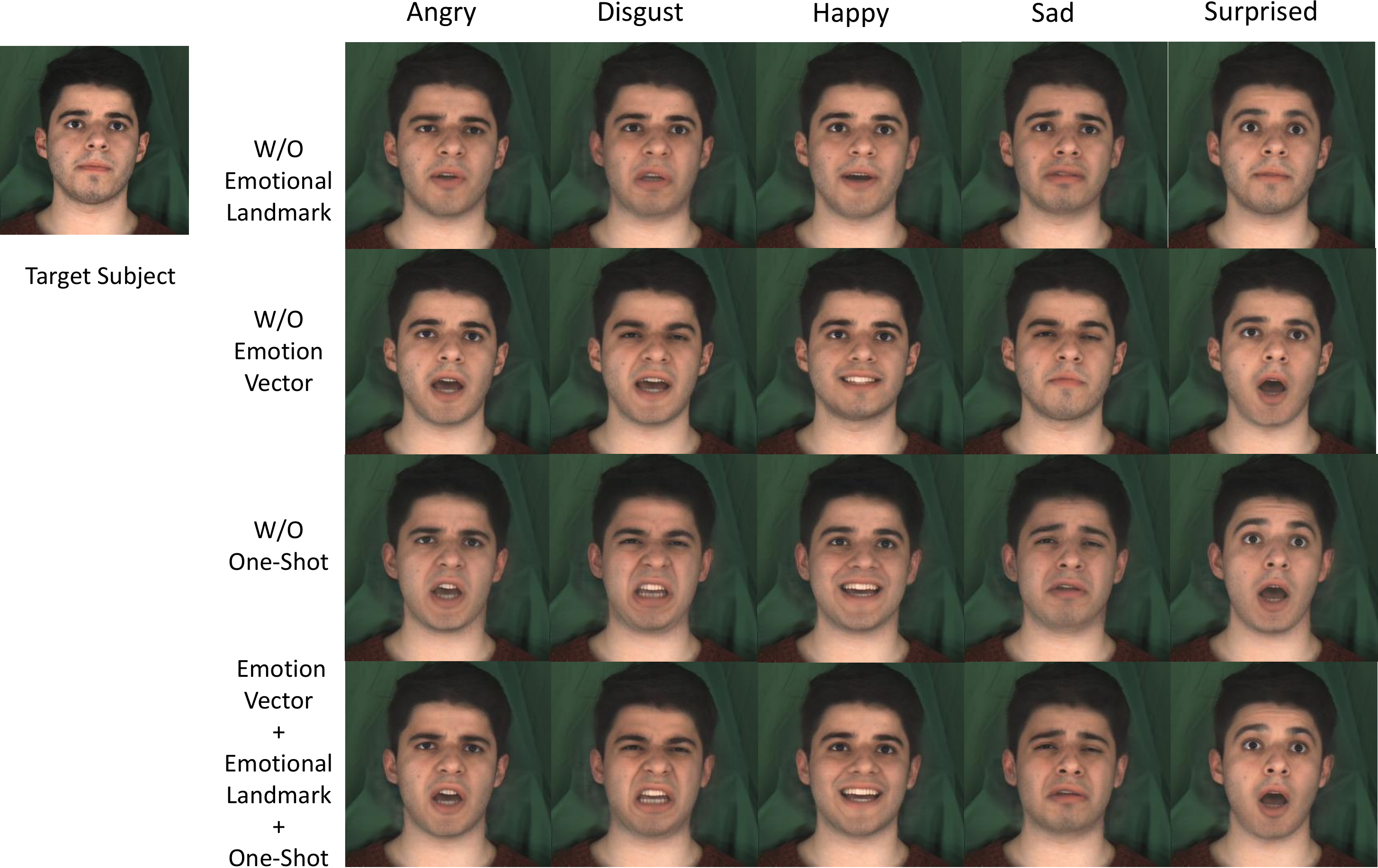}
    \caption{\small{Qualitative Ablation for Texture Generation Network $G_T$.}}
    \label{fig:ablation_texture}
\end{figure}

\textbf{Landmark Generation Network $G_L$:} An ablation study of $G_{L}$ is presented in Table \ref{tab:ablation_landmark}. 
 \textit{(1) Ours w/o Graph Encoder} is a variation of our network $G_L$ with only Audio Encoder $E_A$, Emotion Encoder $E_E$ and Graph Decoder $D_G$.
\textit{(2) Ours w/o skip connections} is without skip connections between Graph Encoder $E_G$ and Graph Decoder $D_G$ (shown Fig. \ref{fig:blockdiagram}). 
\textit{(3) Ours w/o edge weights} is without using the learnable edge weights $\boldsymbol{\omega}$ in Eqn. \ref{eqn:graphcnn}.
\textit{(4) Ours w/o ${L_{gan}}$} is without adversarial learning. 
As Table \ref{tab:ablation_landmark} demonstrates, our proposed network in Fig. \ref{fig:blockdiagram} trained with the losses in Eqn. \ref{eqn:loss_landmark} leads to improved results. 
\\
\textbf{Texture Generation Network $G_{T}$:} An ablation study of $G_{T}$ is presented in Table \ref{tab:ablation_texture}. %Please refer \textbf{\textit{TA (Sec. 2.7 and Fig. 3)}} for qualitative results.
\textit{(1) Ours w/o emotion feature:} Without the concatenated emotion feature input $f_e$, the emotion accuracy highly degrades (Table \ref{tab:ablation_texture}) as the network cannot generate frowns (for disgust, angry) or eyebrow-raising (for happy, surprise), or lowering (for sad) from emotional landmarks only, as shown in Fig. \ref{fig:ablation_texture} (second row). As CSIM is calculated between the predicted frame and the input neutral identity face of the target, the value of CSIM without emotion feature is higher. \textit{(2) Ours w/o emotional landmark:} When the texture is generated from only speech-induced landmarks (without emotion) the emotion accuracy decreases. Learning emotion on landmarks helps generate facial expressions especially in the mouth region for emotions like happy, angry, sad, and disgust. Fig. \ref{fig:ablation_texture} (top row) shows that without emotional landmark, emotion rendering is very restricted. 
\textit{(3) Ours w/o one-shot learning:} One-shot learning helps to achieve better identity preservation.  As can be seen in Fig. \ref{fig:ablation_texture} (last row) the facial structure, skin color of the target subject are better captured in our final animation with one-shot learning.

\subsection{User Study:}
\begin{figure}[t]
  \centering
  %\vspace{4mm}
  %\includegraphics[width=\columnwidth]{LaTeX/TextureStage-crop.pdf}
  \includegraphics[width=0.9\columnwidth,left]{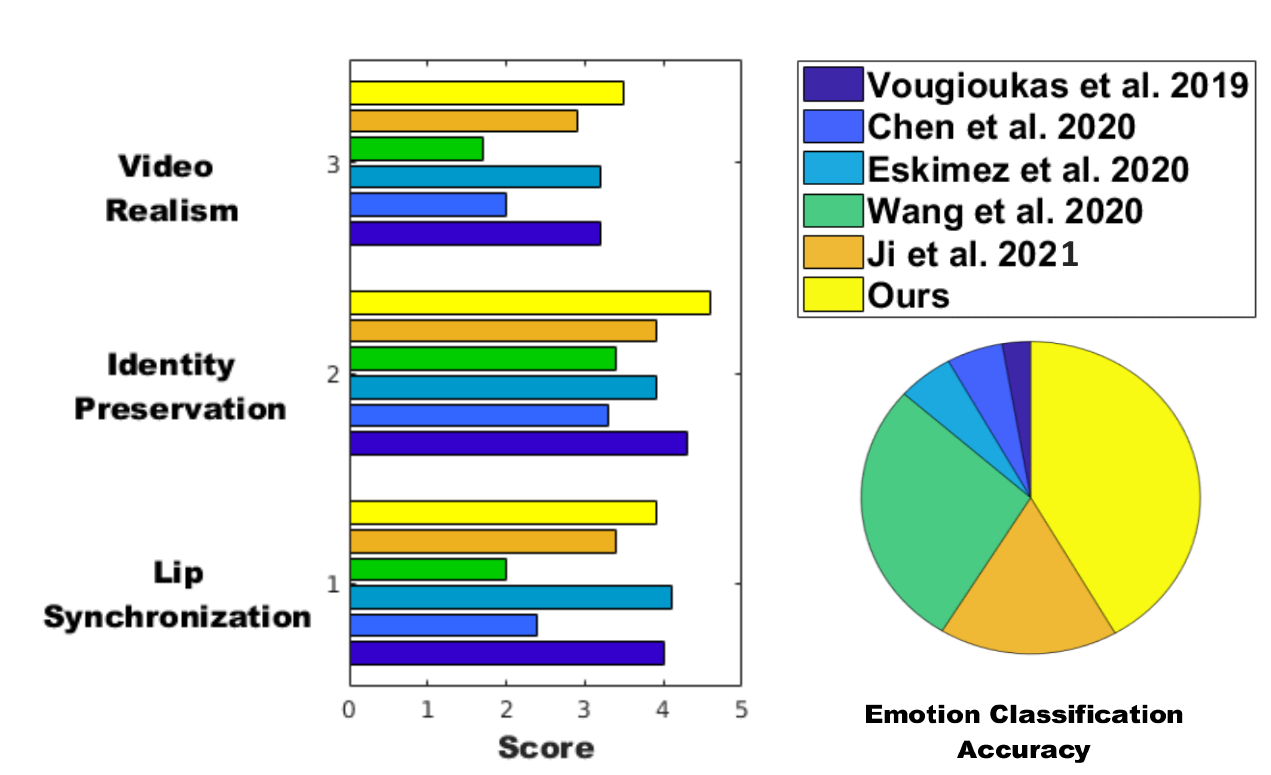}
  \caption{ \small{User Study results. The bar plots represent the average score (range 0-5, high score indicates better performance). 
  }}
  \label{fig:userstudy_emo}
  \end{figure}

We have conducted a user study for subjective evaluation of our method against SOTA. $26$ participants rate total $30$ videos from  \cite{vougioukas2019realistic,eskimez2020speech,chen2020talking}, MEAD, EVP and our method. Each video is evaluated for lip sync, identity preservation, and video realism. Additionally, the participants also classify the emotion perceived from the video. 
%The videos are evaluated on test data from MEAD and CREMA-D datasets on which of the SOTA methods are trained.
The results are shown in Fig. \ref{fig:userstudy_emo}. 
Overall our method achieves comparable performance in lip-sync and better performance over SOTA methods in identity preservation, emotion classification accuracy, and realism in video generation.  

\section{Conclusion}
We propose a speech-driven emotion-controllable generalized emotional talking face generation method that uses a single image of an arbitrary target person in neutral emotion to generate animation in different emotions. We use Graph convolution for geometry-aware motion and emotion generation on facial landmarks. With one-shot learning, our emotion-guided optical flow-based texture deformation network can generalize better for arbitrary target subjects when compared to existing SOTA methods. Our animation results on different benchmark datasets and for different celebrity faces show more realistic animation than SOTA methods. However, our method currently synthesizes fixed head poses. In future work, audio and emotion-driven head movements can be added for enhanced realism of emotional talking face animation. %Also, the fixed-background images from existing emotional audio-visual datasets can be augmented with more generalized synthetic backgrounds for training, in order to eliminate the background replacement operation.

%% The file named.bst is a bibliography style file for BibTeX 0.99c
\bibliographystyle{named}
\bibliography{ijcai22}

\appendix
\section{Appendix}	

%\subsection{\textbf{Spontaneous Eye Blink Generation on Facial Landmarks}}
%Eye blinks are essential for realism of synthesized face animation, but this is independent of speech. Therefore, we adapt an unsupervised method for realistic eye blinks generation as proposed in \cite{dasspeech} from random noise input $\mathcal{N}(\mu,\,\sigma^{2}) $.
%The blink generator network learns the blink pattern and duration of blinks and generates a sequence of eye landmark displacements on the canonical face by minimizing the MMD (Maximum Mean Discrepancy) \cite{gretton2007kernel} loss. MMD loss is defined defined as follows:
%%\textbf{Objective Function for blink prediction}
%\begin{multline}
%\mathcal{L}_{MMD}=\mathbb{E}_{(X+\delta),(X+\delta)^\prime \sim p}\mathcal{K}((X+\delta),(X+\delta)^\prime) \\ + \mathbb{E}_{\hat{X},\hat{X}^\prime \sim q}\mathcal{K}(\hat{X},\hat{X}^\prime) - 2\mathbb{E}_{(X+\delta) \sim p, \hat{X} \sim q}\mathcal{K}((X+\delta),\hat{X})
%\label{equ:mmd_loss}
%\end{multline}
%\noindent
%where, 
%%$X$,$Y$ are the distributions of ground-truth and predicted eye landmark displacements over the sequence respectively; 
%$\mathcal{K}(x,y)$ is defined as $exp(-\frac{|x-y|^2}{2\sigma})$, $p$ and $q$ represents samples from distributions predicted, $X+\delta$ and GT landmarks, $\hat{X}$. $\delta$ is the predicted landmark displacements.

\subsection{Experiments and Training Details}
\subsubsection{Dataset Description}

The recently introduced MEAD dataset \cite{wang2020mead} contains sentences recorded by 48 subjects with neutral and 7 different emotions at three different intensities.  CREMA-D \cite{cao2014crema} contains 7,442 audio-visual clips of 91 actors from different ethnic backgrounds with 10 sentences uttered at 5 different emotions at 3 different intensities. RAVDESS \cite{livingstone2018ryerson} dataset contains two sentences uttered with 7 emotions at two intensity levels by 24 professional actors.
We use MEAD dataset for training and evaluation. CREMA-D and RAVDESS are used for evaluation only

\subsubsection{Data Pre-Processing} \label{sec:supp-datapre}
\begin{figure}[ht]
\centering
\includegraphics[width=\columnwidth]{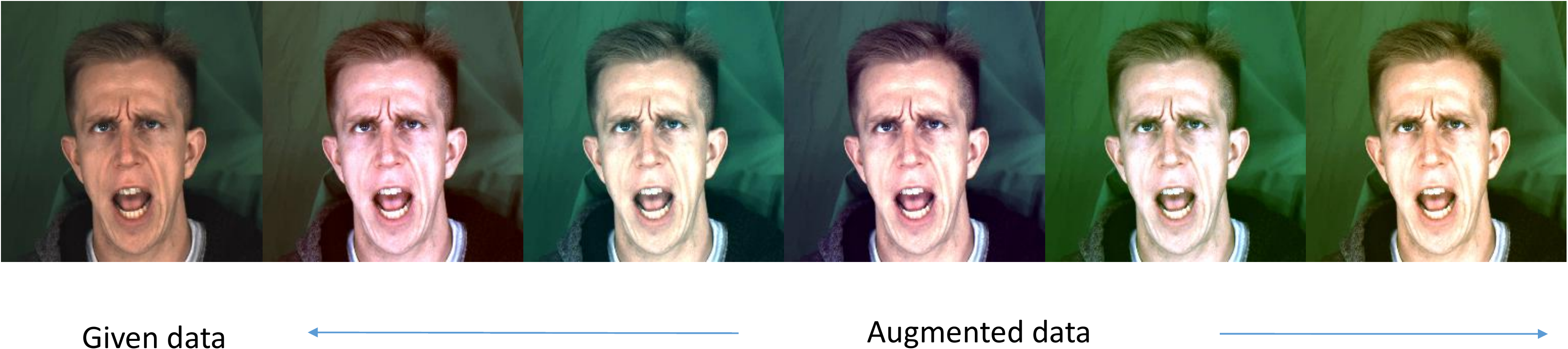}
\caption{Data augmentation}
\label{fig:data_aug}
\end{figure}

Training landmarks are detected using a combination of 3D landmarks from \cite{guo2020towards} and face parsing \cite{yu2018bisenet} for accurate mouth shapes. GT landmarks are aligned and retargeted to a neutral frontal canonical landmark (similar to \cite{dasspeech}) for training landmark generation network $G_{L}$ . The predicted landmarks from $G_{L}$ are retargeted to target-specific landmarks using an inverse process (\cite{dasspeech}).
The face videos of MEAD dataset are cropped and resized to $256$x$256$ and warped to a frontal pose for training texture generation network $G_{T}$. MEAD dataset \cite{wang2020mead} has limited variety in color and illumination. To overcome the challenges in generalization for large number of out of distribution samples, we perform a data augmentation (Fig. \ref{fig:data_aug}) with the MEAD dataset during training.

\subsubsection{Network Architectures} \label{sec:supp-networkarch}

\textit{\textbf{Speech and Emotion Driven Landmark Generation Network $G_L$ :}} The Audio Encoder network $E_A$ encodes DeepSpeech features  $\mathcal{A} = \{a_t \in \mathbb{R}^{6 \times 29}\}$ for each video frame at time $t$. DeepSpeech \cite{huang2017arbitrary} features $\{a_t \in R^{W \times 29}\}$ are extracted for a temporal sliding window of size $W=6$ centered at each video frame. $E_A$ consists of $3$ LSTM layers which encode input size $29$ to a hidden size $256$. The LSTM network output is mapped to a feature $\mathbf{f_a}$ of size $128$ for each $a_t$. 
\\
The Emotion Encoder $E_E$ consists of a single convolutional layer to map emotion vector input $e$ (concatenation of one-hot vectors for emotion type and intensity) to $128$-dimensional encoder feature $\mathbf{f_e}$.
\\
Hierarchical Encoding of Graph : The face landmark graph $\mathcal{G}= (\mathcal{V}, \mathcal{E}, A)$ is first divided into $K=8$ subsets of vertices, each representing a facial region, e.g., eye, nose, etc. Graph pooling is performed on each vertex set $\{V_{i}|i=1,2 \cdots K\}$ to generate features of a smaller graph $G^{1}$ with $K$ vertices. Further graph convolution and pooling are done to create a $d=128$ dimensional feature $ \mathbf{f_l} \in R^d $ representing a graph $G^{2}$ consisting of a single vertex representing the entire face. 
\\
The Graph Encoder network ${E}_{G}$ consists of 3 graph convolutional (GCN) layers , followed by Pooling, to map input graph of size $64 \times 2$ to pooled graph $G^{1}$ with feature size $8 \times 128$. This is again followed by graph convolution and pooling to a graph $G^{2}$ with feature size $1 \times 128$, representing the encoded feature $\mathbf{f_l}$ of size $128$. 
\\
Graph Decoder $D_G$ performs graph upsampling in order to reconstruct the final output graph. Graph convolution intermediate features in $E_G$ are added via skip connections during upsampling in ${D}_G$ for preserving facial geometry-related information during graph reconstruction. Graph upsampling in Graph Decoder $D_G$ uses skip connections from ${E}_{G}$.
\\
The architecture of Graph Discriminator $D_L$ is similar to the Graph Encoder network ${E}_{G}$ as it uses graph convolution and pooling, but the output of $D_{L}$ is a realism score that determines how real or fake is the graph $\mathcal{G'}$ generated by $G_L$.

\begin{table*}[tb]
%\captionsetup{font=scriptsize }
\centering
    \noindent\resizebox{0.6\textwidth}{!}{
\begin{tabular}{|c|c|c|c|c|c|c|c|c|c|c|}
\hline 
 Method & PSNR & SSIM & CPBD & FID & M-LD & M-LVD & F-LD & F-LVD  & CSIM & $Sync_{conf}$\\ 
\hline 
 MakeItTalk & 24.89 & 0.77 & 0.219 & 158.76 & 7.65 & 0.59 & 5.44 & 0.46 & 0.57 & 3.22 \\
 Wav2Lip & 25.37 & 0.79 & 0.282 & 127.22 & 6.74 & \textbf{0.39} & 4.91 & \textbf{0.26} & 0.84 & \textbf{6.28} \\
 Ours & \textbf{28.78} & \textbf{0.80} & \textbf{0.385} & \textbf{30.41} & \textbf{1.53} & 0.49 & \textbf{0.97} & 0.34 & \textbf{0.91} & 3.27 \\
\hline
\end{tabular}
}
\caption{\small {Results on neutral emotion of MEAD dataset.}}
\label{tab:quantitativenoem}
\end{table*}

\textit{\textbf{Texture Generation Network $G_T$ :}}
Image encoder network $E_T$ consist of convolutional layers of size $256\times256\times64, 128\times128\times128, 64\times64\times256$ respectively. The heatmap difference is encoded as $64\times64\times69$ (68 facial landmarks + background). The concatenated image feature and heatmap difference ($325$ channels) is passed through an encoder with downsampling blocks of sizes $64\times64\times512, 32\times32\times1024, 16\times16\times1024$. Emotion encoder feature is added to this layer and passed through a decoder with upsampling blocks of size $16\times16\times1024,32\times32\times512, 64\times64\times256$. Softmax and Tanh activaions are applied at this layer to predict occlusion map and optical flow map respectively. The concatenated occlusion and flow maps are passed through decoder. The convolutional layers of the decoder are $64\times64\times256,128\times128\times128,256\times256\times64,256\times256\times3$. Adaptive Instance Normalization \cite{huang2017arbitrary} is used at the bottleneck layer of the decoder $D_T$. Output feature of $E_T$ is used for layerwise instance normalization of bottleneck layers of $D_T$. At test time we fine-tune $G_T$ with a single neutral target face and update only the weights of $E_T$ and $D_T$ keeping the rest of the network weights of $G_T$ fixed.  %Frame Discriminator $D$ has similar architecture as $E_T$. 
\begin{figure}[tb]
  \centering
  \includegraphics[width=0.9\columnwidth,trim={0cm 0cm 0 0cm},clip]{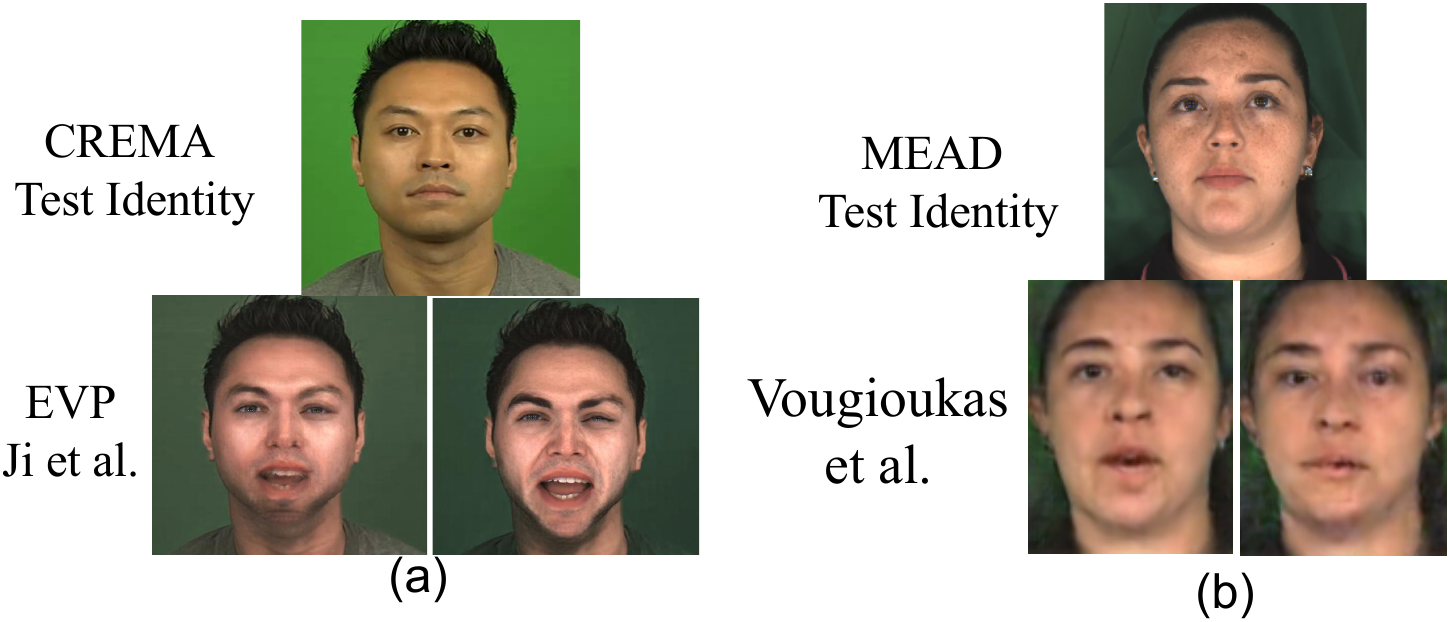}
  \caption{ \small{(a) Evaluating EVP \protect \cite{ji2021audio} on CREMA-D test subject. (b) Evaluating  \protect \cite{vougioukas2019realistic} on MEAD test subject.}}
  \label{fig:suppfig1}
  \end{figure}
 \label{sec:supp-evp-cr}

\subsubsection{Results on Neutral Emotion}

We present quantitative results of some recent talking face generation methods that do not render emotion - Wav2Lip \cite{prajwal2020lip}, MakeItTalk \cite{zhou2021pose} in Table \ref{tab:quantitativenoem}. For a fair comparison we evaluate only for the neutral emotion videos. Although Wav2Lip produces better lip synchronization on neutral emotion videos, our method is capable of rendering facial emotions, which is a key element of realistic facial animation.

\subsubsection{Generalization failure of SOTA methods} \label{sec:supp-evp-cr} 
The emotional talking face SOTA methods MEAD \cite{wang2020mead} \footnote{\label{fn:mead}https://github.com/uniBruce/Mead}, EVP \footnote{\label{fn:evp}https://github.com/jixinya/EVP} \cite{ji2021audio} have publicly available pretrained texture models that are subject-specific. %Morever these methods \footref{fn:mead} \footref{fn:evp} use 106 landmarks from a landmark detection algorithm which is not open source, hence fine-tuning on CREMA subjects is difficult. 
Evaluating EVP using the publicly available pre-trained model of a MEAD subject results in texture generation failure for a test subject from CREMA-D (shown in Fig. \ref{fig:suppfig1}(a)). Hence for fairness of comparison we have not evaluated MEAD \cite{wang2020mead}, EVP \cite{ji2021audio} on CREMA-D dataset for qualitative and quantitative comparison.

The publicly available pre-trained models of  \cite{vougioukas2019realistic,eskimez2020speech} are trained on CREMA-D dataset. These methods fail to generalize to MEAD subjects (an example shown in Fig. \ref{fig:suppfig1} (b)). Hence for a fair comparison we do not evaluate \cite{vougioukas2019realistic,eskimez2020speech} on MEAD for qualitative and quantitative comparison.

\subsubsection{One-shot Learning on Emotional Talking Face} \label{sec:supp-oneshot}

\begin{figure}[t]
  \centering
  \includegraphics[width=0.5\columnwidth,trim={0cm 0cm 0 0cm},clip]{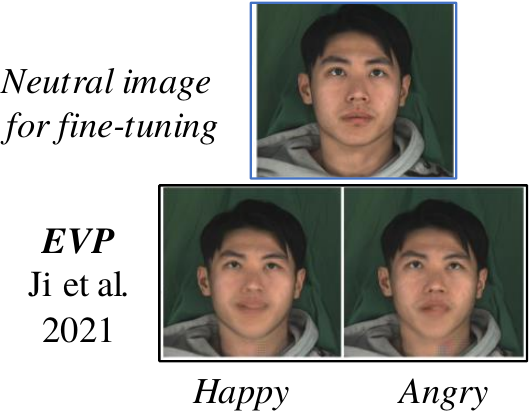}
  \caption{ \small{ One-shot fine-tuning results of EVP \protect \cite{ji2021audio}.}}
  \label{fig:suppfig3}
  \end{figure}
  
To emphasize upon the advantage of our one-shot fine-tuning with respect to other emotional talking face SOTA methods, we fine-tune the Edge-to-Video Translation network of EVP \cite{ji2021audio} on a single image of a target in neutral emotion and fixed headpose. The evaluation result of EVP (shown in Fig. \ref{fig:suppfig3}) shows that although the identity is preserved, emotions are not captured well and the mouth region contains texture blur.

\subsubsection{Background Replacement}
\begin{figure}[t]
  \centering
  \includegraphics[width=\columnwidth,trim={0cm 0cm 0 0cm},clip]{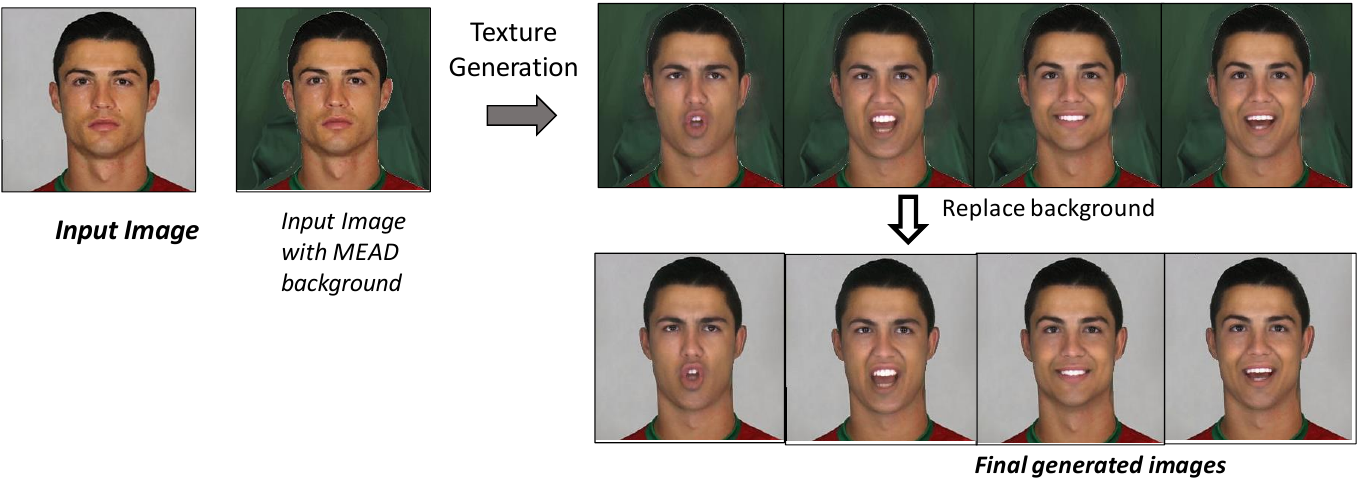}
  \caption{ \small{Background replacement for handling arbitrary backgrounds.}}
  \label{fig:suppfig4}
  \end{figure}

Our texture generation network is trained on MEAD data which contains a fixed dark green background for all subjects. In order to handle different backgrounds at test time, first the background of the given target image is replaced with MEAD background for the texture generation. And as post-processing, the background of the generated images are then substituted with the background of the original input image. This background replacement (Fig. \ref{fig:suppfig4}) helps us process arbitrary unknown target faces using our model pre-trained on MEAD, irrespective of different backgrounds. %Our supplementary video \footnote{https://youtu.be/qqqRx-z5l9k} shows results on arbitrary celebrity faces with different backgrounds. 
In future, to eliminate the background replacement operation, the fixed-background images from existing emotional audio-visual datasets can be augmented with more generalized synthetic backgrounds for training. 

\end{document}